\documentclass[10pt,twocolumn,letterpaper]{article}
\usepackage[T1]{fontenc}

\usepackage[pagenumbers]{cvpr}

\usepackage{booktabs}
\usepackage{multirow}
\usepackage{graphicx}

\definecolor{cvprblue}{rgb}{0.21,0.49,0.74}
\usepackage[pagebackref,breaklinks,colorlinks,allcolors=cvprblue]{hyperref}
\usepackage{array}
\usepackage{xcolor}
\providecommand{\venue}[1]{\textcolor{gray}{\scriptsize #1}}

\title{IGGLoop: Instance-Grounded Gaussian Looped Transformer \\
for Feed-Forward 4D Driving Scene Reconstruction}

\author{%
Guanfeng Tang$^{1,2*}$ \quad Renkai Luo$^{1,2,3*}$ \quad Cheng Dai$^{1,2}$ \quad Siming Li$^{1,2}$ \quad Hao Zhao$^{1,2\dagger}$\\
{\normalsize $^{1}$AIR, Tsinghua University \quad $^{2}$PrimalVerse \quad $^{3}$Qiuzhen College, Tsinghua University}\\
{\small $^{*}$Equal contribution. \quad $^{\dagger}$Corresponding author.}
}

\begin{document}
\twocolumn[{
    \renewcommand{\twocolumn}[1][]{#1}
    \maketitle
    \vspace{-12pt}
    \begin{minipage}{\textwidth}
        \centering
        \includegraphics[width=0.99\linewidth]{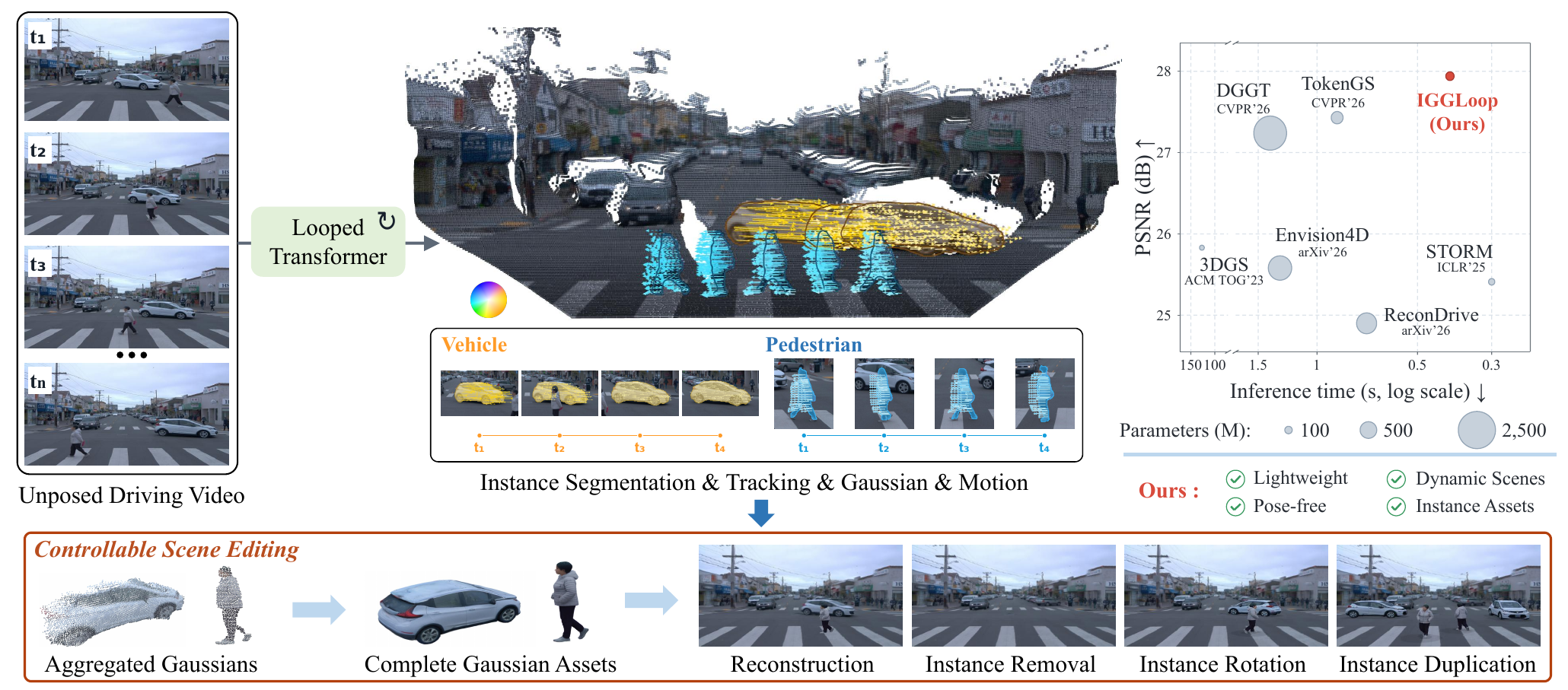}
        \captionof{figure}{
        \textbf{IGGLoop enables instance-grounded 4D scene reconstruction and controllable scene editing.}
        Given unposed driving videos, our framework (upper left) unifies instance segmentation, tracking, Gaussian reconstruction,
        and motion estimation with a compact Looped Transformer. Motion-aligned instance Gaussians are aggregated and completed using a generative prior, enabling controllable scene editing (bottom). The comparison on Waymo (upper right) demonstrates high
        reconstruction quality with efficient inference and a compact model, where bubble size indicates parameter count.
        }
        \label{fig:teaser}
    \end{minipage}
    \par\vspace{12pt}
}]
\begin{abstract}
Reconstructing dynamic driving scenes for controllable re-simulation requires accurate reconstruction of individual traffic participants and their motion, together with the ability to manipulate them independently. Existing feed-forward approaches primarily target scene-level visual fidelity, while instance-level modeling and control often rely on auxiliary models or 3D annotations. Our key observation is that persistent instance associations allow complementary visual cues to be aggregated across frames for joint Gaussian reconstruction and motion estimation. Therefore, we present IGGLoop, an instance-grounded framework for feed-forward 4D driving scene reconstruction that unifies instance segmentation, tracking, Gaussian reconstruction, and motion estimation. Instance queries predict object masks and maintain identities across frames, enabling Gaussian and motion queries to aggregate complementary observations of each instance. A shared-weight Looped Transformer iteratively refines multi-frame geometric features for camera pose and depth estimation. Experiments on Waymo, nuScenes, and Argoverse~2 demonstrate competitive reconstruction quality, with a 2.57\% PSNR gain over the state-of-the-art pose-free baseline on Waymo, only 6.4\% of its parameters, and a 3.45$\times$ speedup. Using the predicted instance identities and motion, we further aggregate cross-frame observations and recover complete Gaussian assets while preserving their scene poses. These assets support independent instance insertion, removal, and repositioning for controllable re-simulation.
\end{abstract}

\section{Introduction}
\label{sec:intro}
Dynamic scene reconstruction and re-simulation of real-world driving scenes are essential for training and evaluating autonomous driving systems~\cite{neurad,unisim,splatad,freesim,rad,kim2021drivegan,lu2025infinicube}, with reconstruction methods evolving from iterative per-scene optimization~\cite{3dgs,deformablegs,emernerf,pvg,streetgaussians,drivinggaussian,beziergs,rematas2022urbanradiancefields,niemeyer2022regnerf,yu2024mipsplatting,huang2024_2dgs,wu2024reconfusion} toward feed-forward prediction~\cite{gs_lrm,noposplat,storm,vggt,tokenGS,ReconDrive,dynamicvggt,envision4D,dggt,drivingforward,drivingrecon,mcsplat,zhou2018stereomagnification,tucker2020singleviewmpi,flynn2019deepview,jin2025lvsm}. In driving scenarios, instances such as vehicles and pedestrians are major sources of dynamics, making them central to scene editing and controllable re-simulation~\cite{assetharvester,cadsim,mars,omnire,horizonforge,real2sim,ost2021neuralscenegraphs,fischer2024multilevelnsg,kundu2022panopticneuralfields}. Such applications require the ability to extract, complete, and reposition the 3D Gaussian representations of individual instances while preserving the surrounding environment~\cite{completevehicle}.

However, existing methods primarily focus on scene-level reconstruction, overlooking the need for explicit instance modeling in autonomous driving simulation. Token-aligned approaches such as TokenGS~\cite{tokenGS} initialize a fixed number of learnable Gaussian tokens and decode each into a group of Gaussians representing an arbitrary scene region, thereby impeding the direct extraction of instance-specific Gaussian assets. Pixel-aligned methods such as DGGT~\cite{dggt} assign a distinct Gaussian to each image pixel, providing a basis for instance-level Gaussian extraction. However, existing approaches still rely on 2D dynamic mask supervision to identify moving pixels, off-the-shelf 3D point tracking models for motion estimation~\cite{TAPIP3D}, and external video segmentation models to distinguish and track individual instances. This reliance on multiple components significantly increases computational overhead and allows errors to propagate and accumulate across stages~\cite{streetgaussians_notr}. These limitations motivate us to develop an end-to-end instance-grounded framework that jointly learns instance segmentation, tracking, Gaussian reconstruction, and motion estimation within a unified framework.

Our key insight is that partial observations of the same instance across frames provide complementary geometric and appearance cues, while their temporal variation informs motion estimation. Therefore, we introduce IGGLoop, a pose-free feed-forward framework that coordinates instance, Gaussian, and motion queries for 4D driving scene reconstruction. Specifically, we adopt a Looped Transformer~\cite{deja_view} that iteratively refines multi-frame geometric features through a shared set of Transformer blocks, enabling effective camera pose and depth estimation with a compact backbone. We subsequently initialize a set of learnable instance queries to aggregate object-specific visual information for mask prediction and propagate their updated representations across frames to maintain consistent object identities. Guided by these associations, a set of Gaussian queries aggregates complementary features from the same instance across frames into contexts that condition dense predictions of Gaussian geometry and appearance. In parallel, a separate set of motion queries captures instance-specific temporal dynamics to condition dense 3D motion prediction at target timestamps. To stabilize motion learning under image reconstruction supervision, we further introduce a depth validity loss that discourages predicted displacements from placing Gaussians behind the target camera.
Based on the resulting Gaussians, instance identities, and motion estimates, we further recover complete Gaussian assets using a generative prior while preserving their original scene poses.

On Waymo~\cite{waymo}, our IGGLoop achieves 27.94 dB PSNR and 0.826 SSIM with only 176.52M parameters, outperforming the compared optimization-based and feed-forward methods in both metrics. Compared with the state-of-the-art (SOTA) pose-free feed-forward reconstruction method DGGT~\cite{dggt}, it improves PSNR by 0.70 dB while using only 6.4\% of its parameters and achieving a $3.45\times$ speedup. The framework also demonstrates strong zero-shot generalization, achieving 27.27 dB PSNR on nuScenes~\cite{nuscenes} and 26.25 dB on Argoverse~2~\cite{argoverse2}, with the best PSNR, SSIM, and LPIPS among the compared methods on both datasets. Additionally, the recovered complete Gaussian assets support autonomous driving simulation by enabling flexible instance-level scene editing, including translating and rotating individual vehicles and pedestrians.

\section{Related Work}
\label{sec:related}
\subsection{Feed-forward Reconstruction}
Feed-forward reconstruction predicts scene representations, such as neural radiance fields~\cite{pixelnerf,ibrnet,mvsnerf} or 3D Gaussians~\cite{chen2024lara,ye2026yonosplat}, from input observations in a single forward pass, enabling efficient reconstruction of unseen scenes without scene-specific fitting~\cite{lgm,flash3d,occsplat}. GS-LRM~\cite{gs_lrm}, MVSplat~\cite{mvsplat}, NoPoSplat~\cite{noposplat}, and DepthSplat~\cite{depthsplat} establish this paradigm for static scenes~\cite{pixelsplat,splatt3r,pansplat,latentsplat,freesplat,xu2026resplat,chen2024mvsplat360}. VGGT~\cite{vggt} further unifies camera estimation and dense geometry prediction, providing geometric priors for feed-forward reconstruction frameworks such as DGGT~\cite{dggt} and DynamicVGGT~\cite{dynamicvggt}. However, its large transformer architecture incurs substantial parameter and GPU memory costs~\cite{fastvggt,goodtokenhunting,qvggt,litevggt,regimevggt}. To enable more efficient dynamic scene reconstruction, we adopt a shared-weight Looped Transformer~\cite{deja_view}, enabling iterative geometric refinement within a compact framework while supporting an arbitrary number of input frames without requiring camera poses.

\subsection{Dynamic Scene Reconstruction}
Dynamic scene reconstruction recovers time-varying geometry and appearance from image sequences~\cite{dnerf,nerfies,nsff,tensor4d,dynamic3dgs,four_dgs,coda4dgs,adgs,li2023dynibar,turki2023suds,fridovichkeil2023kplanes,cao2023hexplane,zhang2025monst3r}. Recent feed-forward approaches differ in how they organize scene dynamics~\cite{four_dgt,l4gm,pointforwardv2,instantnurec,fruc,drive2gauss,lgs,unisplat}. One line adopts mask- or box-guided static-dynamic composition. DGGT~\cite{dggt} combines supervised dynamic mask prediction with a pretrained 3D point tracking module. ReconDrive~\cite{ReconDrive} combines SAM2~\cite{SAM2} masks with rigid motion derived from 3D box annotations. PointForward~\cite{pointforward} uses temporally annotated 3D boxes to assign dynamic queries, canonicalize instances, and propagate motion. These approaches rely on external motion estimation components or 3D bounding-box annotations to model object dynamics, introducing additional model or annotation requirements. Another line models temporal evolution directly through learned Gaussian attributes or deformations. STORM~\cite{storm} predicts Gaussian velocities, while Envision4D~\cite{envision4D} learns time-conditioned nonlinear motion. TokenGS~\cite{tokenGS} represents dynamic scenes with static and time-conditioned dynamic Gaussian tokens, while DynamicVGGT~\cite{dynamicvggt} predicts Gaussian velocities under scene-flow supervision. However, these temporal representations do not explicitly organize observations into persistent instance-specific contexts for geometry and motion reconstruction. In contrast, our instance queries organize cross-frame evidence for Gaussian and local motion prediction, enabling motion-aligned aggregation and generative completion of instance assets.

\section{Method}
\label{sec:method}

\begin{figure*}
    \centering
    \includegraphics[width=0.99\linewidth]{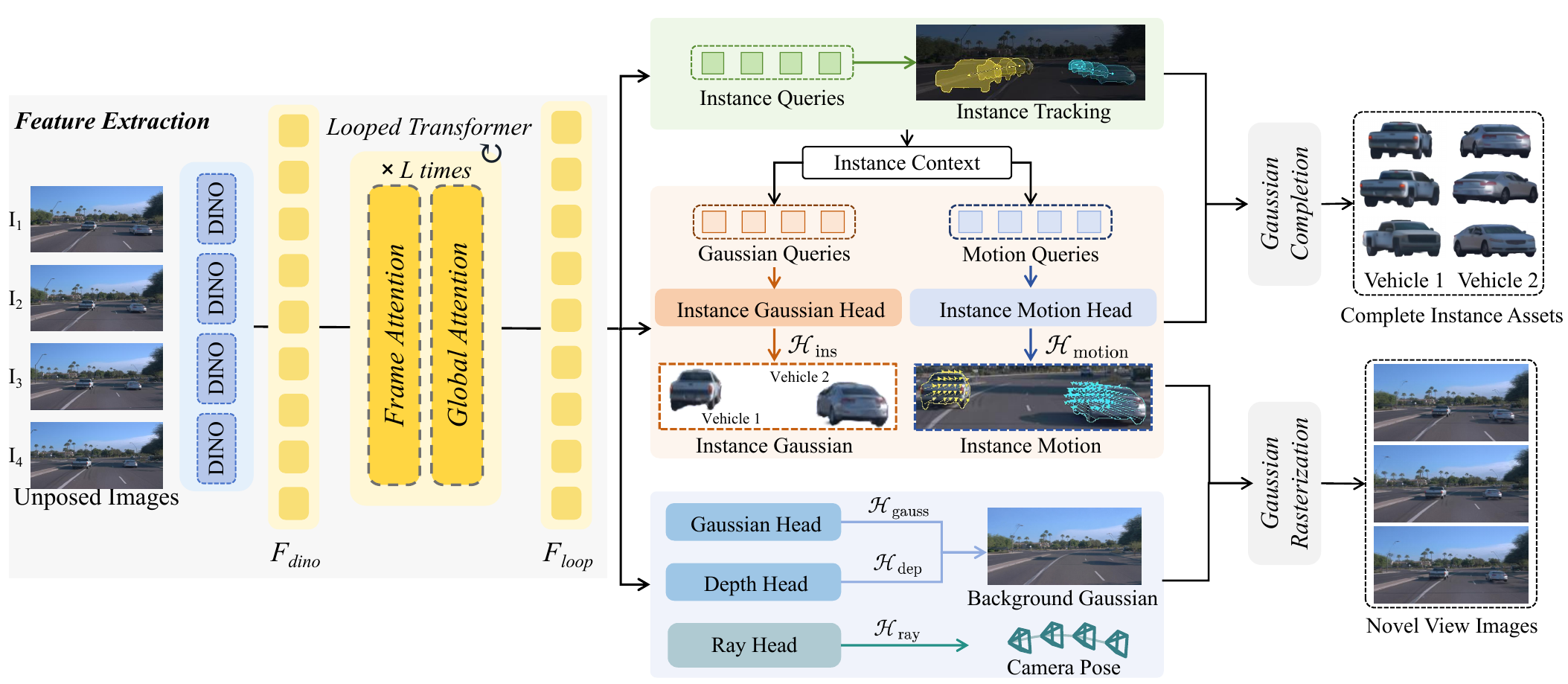}
    \caption{
    \textbf{Overall architecture of IGGLoop.}
    Given unposed images of a dynamic driving scene, a DINOv2 encoder and a shared-weight Looped Transformer extract semantic and geometric features, respectively.
    Instance queries predict masks and maintain cross-frame identities, guiding Gaussian and motion queries to aggregate instance-specific context for Gaussian reconstruction and motion estimation.
    In parallel, depth, ray, and Gaussian heads estimate scene geometry, camera poses, and background Gaussians.
    The predicted Gaussians and motion further support novel-view synthesis through Gaussian rasterization and instance asset completion through a generative prior.
    }
    \label{fig:framework}
\end{figure*}

\begin{figure}
    \centering
    \includegraphics[width=0.99\linewidth]{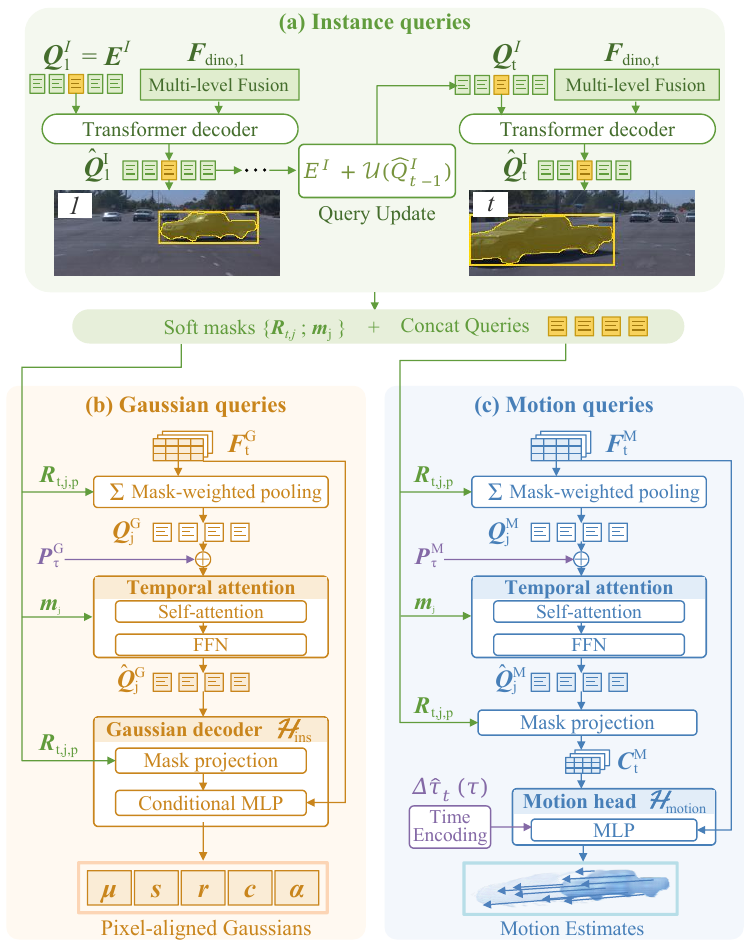}
    \caption{\textbf{Query architecture of IGGLoop.}
    Instance queries establish cross-frame associations, guiding Gaussian and motion queries to aggregate instance-specific features for reconstruction and motion estimation.}
    \label{fig:query}
\end{figure}
We first present the architecture of IGGLoop, a pose-free feed-forward framework for 4D Gaussian reconstruction built on a Looped Transformer backbone (Sec.~\ref{architecture}).
We then describe our instance-grounded query modeling, including instance queries for segmentation and tracking, Gaussian queries for reconstruction, and motion queries for motion estimation (Sec.~\ref{sec:query-model}).
Finally, we explain how the predicted instance identities, Gaussians, and motion enable us to aggregate observations across frames to support instance asset completion and scene editing (Sec.~\ref{sec:completion}).

\subsection{Model Architecture}
\label{architecture}

\paragraph{Semantic and Geometric Feature Extraction}
As illustrated in Fig.~\ref{fig:framework}, IGGLoop takes as input a sequence of unposed RGB images $\{I_t\}_{t=1}^{T}$ captured at timestamps $\{\tau_t\}_{t=1}^{T}$, with a temporal span $T_{\mathrm{seq}}$.
To support instance-level scene understanding~\cite{siddiqui2023panopticlifting}, we first extract multi-level semantic features $\mathbf{F}_{\mathrm{dino},t}$ from each image $I_t$ using a pretrained DINOv2 encoder~\cite{dinov2}.
We collectively denote these features by $\mathbf{F}_{\mathrm{dino}}=\{\mathbf{F}_{\mathrm{dino},t}\}_{t=1}^{T}$.
Reconstructing the scene in 3D further requires reasoning about spatial relationships across input frames~\cite{schonberger2016pixelwise,duzceker2021deepvideomvs,wang2024dust3r,leroy2024mast3r,wang2025cut3r}, which we achieve by jointly processing the final-layer image tokens in $\mathbf{F}_{\mathrm{dino}}$ across frames.
Previous feed-forward reconstruction transformers, such as VGGT~\cite{vggt} and VGGT-$\Omega$~\cite{vggt_omega}, rely on deep stacks of independently parameterized layers for multi-view geometric reasoning, resulting in a large number of backbone parameters.
However, geometric predictions are progressively refined through successive Transformer layers, suggesting that part of this computation serves repeated geometric updates~\cite{deja_view}.
Therefore, we adopt a shared-weight Looped Transformer to support iterative geometric refinement.
Specifically, the final-layer image tokens initialize a multi-frame latent state $\mathbf{Z}_0$, which we iteratively refine using a shared Looped Transformer block:
\begin{equation}
    \mathbf{Z}_{k+1}
    =
    \operatorname{Loop}_{\theta}
    \left(\mathbf{Z}_k, s_k, s_{k+1}\right),
    \label{eq:looped_feature_refinement}
\end{equation}
where $k=0,\ldots,L-1$ indexes the refinement iterations.
Here, $\operatorname{Loop}_{\theta}$ performs one refinement step using alternating frame-wise and global self-attention~\cite{vggt}, with the same parameters $\theta$ shared across iterations.
The variables $s_k,s_{k+1}\in[0,1]$ encode the normalized refinement progress at the current and next steps. After $L$ iterations, we extract the geometric features $\mathbf{F}_{\mathrm{loop}}=\{\mathbf{F}_{\mathrm{loop},t}\}_{t=1}^{T}$ from the final state $\mathbf{Z}_L$, where $\mathbf{F}_{\mathrm{loop},t}$ contains the features associated with source frame $t$.
Based on these features, a depth head $\mathcal{H}_{\mathrm{dep}}$ predicts dense depth maps $D_t$, while a ray head $\mathcal{H}_{\mathrm{ray}}$ predicts ray fields from which camera parameters $\Pi_t$ are recovered.

\paragraph{Time-Conditioned Gaussian Representation}
We represent the dynamic scene with pixel-aligned Gaussian maps $\{\mathcal{G}_t\}_{t=1}^{T}$, providing an explicit 3D representation for differentiable rendering.
For each source frame $t$, the map $\mathcal{G}_t=\{G_{t,p}\}_p$ associates each pixel $p$ with a 3D Gaussian parameterized as follows:
\begin{equation}
    G_{t,p}
    =
    \left(
        \boldsymbol{\mu}_{t,p},
        \mathbf{s}_{t,p},
        \mathbf{r}_{t,p},
        \mathbf{c}_{t,p},
        \alpha_{t,p}
    \right),
    \label{eq:gaussian_representation}
\end{equation}
where $\boldsymbol{\mu}_{t,p}$ denotes the 3D center, $\mathbf{s}_{t,p}$ the scale, $\mathbf{r}_{t,p}$ the rotation quaternion, $\mathbf{c}_{t,p}$ the color, and $\alpha_{t,p}$ the opacity.
Gaussian centers are obtained by back-projecting the predicted depths with learned depth corrections using the estimated camera parameters.
All source Gaussians are expressed in a shared scene coordinate system anchored at the first source camera.

Although this coordinate system accounts for camera motion, object motion and depth estimation errors can still cause geometric misalignment across source frames.
Therefore, at a target time $\tau$, we update the centers of both instance and background Gaussians using predicted displacements $\Delta\boldsymbol{\mu}_{t,p}(\tau)$ and modulate their opacities with fusion weights $w_{t,p}(\tau)$, yielding the target-time Gaussian maps $\mathcal{G}_t(\tau)$. 
These maps are jointly rendered at the estimated target camera $\Pi(\tau)$:
\begin{equation}
    \widehat{I}(\tau)
    =
    \mathcal{R}\!\left(
        \bigcup_{t=1}^{T}\mathcal{G}_t(\tau),
        \Pi(\tau)
    \right),
    \label{eq:target_time_rendering}
\end{equation}
where $\mathcal{R}$ denotes differentiable Gaussian splatting and $\widehat{I}(\tau)$ is the rendered image at time $\tau$.

\subsection{Instance-Grounded Query Modeling}
\label{sec:query-model}

\paragraph{Instance Queries for Segmentation and Tracking}
Existing feed-forward methods often focus on scene-level reconstruction, overlooking complementary appearance cues across views of the same instance and the need for explicit instance modeling.
To aggregate complementary observations for instance-grounded Gaussian reconstruction and motion estimation, we first establish persistent instance associations across source frames. Building on query-based set prediction~\cite{detr,cheng2022mask2former}, we initialize a set of instance query slots with learnable embeddings $\mathbf{E}^{I}$, setting $\mathbf{Q}_1^{I}=\mathbf{E}^{I}$ for the first frame, as illustrated in Fig.~\ref{fig:query}.
For each frame $t$, we linearly project the multi-level semantic features $\mathbf{F}_{\mathrm{dino},t}$ into a shared feature space and fuse them through lightweight MLPs.
A shared Transformer decoder~\cite{pmt} jointly processes the fused features and the input queries $\mathbf{Q}_t^{I}$, producing updated queries $\widehat{\mathbf{Q}}_t^{I}$, class scores, and soft instance masks.
For subsequent frames ($t>1$), we construct the input queries from the previous frame's updated queries:
\begin{equation}
    \mathbf{Q}_t^{I}
    =
    \mathbf{E}^{I}
    +
    \mathcal{U}\!\left(
        \widehat{\mathbf{Q}}_{t-1}^{I}
    \right),
    \label{eq:instance_query_propagation}
\end{equation}
where $\mathcal{U}$ is a learnable linear projection.
This propagation carries instance-specific visual information across frames, providing each query slot with a reference for identifying the same instance~\cite{zeng2022motr,meinhardt2022trackformer}.
The shared decoder then updates these slots using current-frame features, accommodating changes in appearance and position while encouraging temporally consistent instance associations.

\paragraph{Instance Gaussian Queries for Gaussian Reconstruction}
We use the predicted soft instance assignments to construct instance Gaussian queries that aggregate geometric and appearance cues across source frames.
A convolutional adapter maps the geometric features $\mathbf{F}_{\mathrm{loop},t}$ to pixel-aligned Gaussian features $\mathbf{F}_t^{G}$.
For each instance $j$, we concatenate its per-frame Gaussian feature representations into a query sequence:
\begin{equation}
    \mathbf{Q}_j^{G}
    =
    \operatorname{Concat}\!\left(
        \frac{\sum_p R_{1,j,p}\mathbf{F}_{1,p}^{G}}
             {\sum_p R_{1,j,p}+\epsilon},
        \;\ldots,\;
        \frac{\sum_p R_{T,j,p}\mathbf{F}_{T,p}^{G}}
             {\sum_p R_{T,j,p}+\epsilon}
    \right),
    \label{eq:gaussian_query_initialization}
\end{equation}
where $\mathbf{F}_{t,p}^{G}$ is the Gaussian feature at pixel $p$ in frame $t$, and $R_{t,j,p}$ is its soft assignment weight to instance $j$.
For the selected instance slots, these weights are derived from predicted mask probabilities and class confidences, with invalid pixels and missing observations masked out.
The constant $\epsilon>0$ ensures numerical stability, and concatenation is performed along the token dimension.

To integrate complementary cues across the observed views, we refine this sequence through temporal self-attention as follows:
\begin{equation}
    \widehat{\mathbf{Q}}_j^{G}
    =
    \operatorname{Temporal\ Attention}\!\left(
        \mathbf{Q}_j^{G}+\mathbf{P}_{\tau}^{G};
        \mathbf{m}_j
    \right),
    \label{eq:gaussian_temporal_attention}
\end{equation}
where $\mathbf{P}_{\tau}^{G}$ contains linearly projected sinusoidal encodings of the normalized source timestamps, and
the mask $\mathbf{m}_j$ excludes frames without valid observations of instance $j$.

An instance Gaussian decoder $\mathcal{H}_{\mathrm{ins}}$ conditions the local Gaussian features $\mathbf{F}_t^{G}$ on the updated Gaussian queries $\widehat{\mathbf{Q}}_j^{G}$ to predict pixel-aligned Gaussian parameters for each instance.
Here, the updated Gaussian queries supply complementary evidence from observations of the same object, while the local features preserve the spatial detail needed to reconstruct individual Gaussians.
Additionally, a background Gaussian head $\mathcal{H}_{\mathrm{gauss}}$ draws on the local geometric features $\mathbf{F}_t^{G}$ to predict Gaussian attributes for the surrounding scene.
For the selected instance slots, soft assignment weights provide a differentiable link between instance prediction and Gaussian reconstruction, enabling joint end-to-end optimization with rendering supervision.

\paragraph{Instance Motion Queries for Motion Estimation}
Instance Gaussian queries aggregate complementary multi-view cues for object reconstruction, but rendering novel views at intermediate times further requires estimating how object motion changes the spatial locations of source Gaussians~\cite{wang2023omnimotion,rajic2025mvtracker}.
To this end, we construct instance motion queries to capture motion cues from observations of the same object over time and predict the displacements needed to align its Gaussians at the target time.

A separate convolutional adapter projects the geometric features into pixel-aligned motion features $\mathbf{F}_t^{M}$.
We construct and refine motion queries following Eqs.~\ref{eq:gaussian_query_initialization} and~\ref{eq:gaussian_temporal_attention}, using independent parameters for temporal attention.
For each frame, we map the refined motion queries back to pixel locations using the soft assignment weights, yielding a dense motion context $\mathbf{C}_t^{M}$.
Let $\Delta\hat{\tau}_t(\tau)$ denote the normalized source-to-target time difference.
The motion head $\mathcal{H}_{motion}$ combines this context, local motion features, and the source-to-target temporal encoding to predict a dense field of target-conditioned average velocities as follows:
\begin{equation}
    \mathbf{v}_t(\tau)
    =
    \frac{s_{\mathrm{scene}}v_{m}}{T_{\mathrm{seq}}}
    \tanh\!\left[
        \mathcal{H}_{motion}\!\left(
            \mathbf{F}_t^{M},
            \mathbf{C}_t^{M},
            \gamma(\Delta\hat{\tau}_t(\tau))
        \right)
    \right],
    \label{eq:motion_prediction}
\end{equation}
where $\gamma$ denotes sinusoidal temporal encoding, $s_{\mathrm{scene}}$ is the scene scale estimated as the median valid predicted depth, and $v_{m}$ bounds the normalized velocity along each coordinate axis.  
We obtain each Gaussian's displacement by multiplying its predicted average velocity by the difference between the target and source timestamps.
For background regions, we also estimate residual displacements to compensate for geometric inconsistencies caused by depth estimation errors, supporting the alignment of background observations across frames.

To regulate the contribution of each source observation, a lightweight fusion head takes the same inputs as $\mathcal{H}_{motion}$ and predicts a positive temporal bandwidth $\beta_{t,p}(\tau)$ in normalized time.
The corresponding fusion weight is computed as follows:
\begin{equation}
    w_{t,p}(\tau)
    =
    \exp\!\left[
        -\frac{1}{2}
        \left(
            \frac{\Delta\hat{\tau}_t(\tau)}
                 {\beta_{t,p}(\tau)}
        \right)^2
    \right],
    \label{eq:temporal_fusion_weights}
\end{equation}
where $w_{t,p}(\tau)\in[0,1]$ controls the contribution of each source Gaussian to the target-time rendering.
We update Gaussian centers using the predicted displacements and weight their opacities by the fusion weights. Together with the source scales, rotations, and colors, these updated parameters define the target-time Gaussian maps $\mathcal{G}_t(\tau)$, which are jointly rendered according to Eq.~\ref{eq:target_time_rendering}.

\begin{table*}[!tp]
    \centering
    \caption{
        \textbf{Quantitative comparison on the Waymo dataset.}
        Higher PSNR and SSIM, and lower LPIPS indicate better performance. Inference time is reported in seconds unless otherwise specified. For per-scene optimization methods, time denotes per-scene fitting time.
    }
    \label{tab:waymo_comparison}

    \scriptsize
    \setlength{\tabcolsep}{0pt}
    \renewcommand{\arraystretch}{1.12}

    \begin{tabular*}{0.8\textwidth}{
        @{\extracolsep{\fill}}lccccccc@{}
    }
        \toprule
        \multirow{2}{*}{\textbf{Method}}
        & \multicolumn{3}{c}{\textbf{Render quality}}
        & \multirow{2}{*}{\textbf{Inference time (s)}}
        & \multirow{2}{*}{\textbf{Parameters (M)}}
        & \multirow{2}{*}{\textbf{Dynamic}}
        & \multirow{2}{*}{\textbf{Pose-free}} \\
        \cmidrule(lr){2-4}
        & PSNR $\uparrow$
        & SSIM $\uparrow$
        & LPIPS $\downarrow$
        & & & & \\
        \midrule

        \multicolumn{8}{@{}l}{
            \textbf{Per-scene optimization}
        } \\
        3DGS~\cite{3dgs}~\venue{TOG'23}
        & 25.83 & 0.762 & 0.18 & 124.48
        & \textbf{59.00} & $\times$ & $\times$ \\
        DeformableGS~\cite{deformablegs}~\venue{CVPR'24}
        & 26.16 & 0.806 & 0.15 & 139.45
        & \underline{59.51} & $\checkmark$ & $\times$ \\
        \midrule

        \multicolumn{8}{@{}l}{
            \textbf{MLLM-based Agent}
        } \\
        GPT-6 Astra~\cite{gpt6}~\venue{OpenAI'26}
        & 18.57 & 0.478 & 0.41 & 43 min
        & - & $\checkmark$ & $\checkmark$ \\
        \midrule

        \multicolumn{8}{@{}l}{
            \textbf{Feed-forward model}
        } \\
        GS-LRM~\cite{gs_lrm}~\venue{ECCV'24}
        & 23.79 & 0.694 & 0.21 & 1.37
        & 86.68 & $\times$ & $\times$ \\
        NoPoSplat~\cite{noposplat}~\venue{ICLR'25}
        & 24.18 & 0.654 & 0.13 & 0.49
        & 612.00 & $\times$ & $\checkmark$ \\
        STORM~\cite{storm}~\venue{ICLR'25}
        & 25.41 & 0.748 & 0.18 & \textbf{0.30}
        & 100.60 & $\checkmark$ & $\times$ \\
        VGGT++~\cite{vggt}~\venue{CVPR'25}
        & 26.36 & 0.792 & 0.12 & 0.83
        & 1256.86 & $\times$ & $\checkmark$ \\
        ReconDrive~\cite{ReconDrive}~\venue{arXiv'26}
        & 24.90 & 0.715 & 0.17 & 0.71
        & 1030.06 & $\checkmark$ & $\times$ \\
        TokenGS~\cite{tokenGS}~\venue{CVPR'26}
        & \underline{27.43} & \underline{0.807} & 0.26 & 0.87
        & 360.08 & $\checkmark$ & $\times$ \\
        DynamicVGGT~\cite{dynamicvggt}~\venue{CVPR'26}
        & 22.50 & 0.628 & 0.19 & 0.80
        & 1318.50 & $\checkmark$ & $\checkmark$ \\
        Envision4D~\cite{envision4D}~\venue{arXiv'26}
        & 25.58 & 0.727 & \underline{0.12} & 1.29
        & 1379.80 & $\checkmark$ & $\checkmark$ \\
        DGGT~\cite{dggt}~\venue{CVPR'26}
        & 27.24 & 0.796 & 0.14 & 1.38
        & 2752.94 & $\checkmark$ & $\checkmark$ \\
        \midrule

        \textbf{Ours}
        & \textbf{27.94}
        & \textbf{0.826}
        & \textbf{0.12}
        & \underline{0.40}
        & 176.52
        & $\checkmark$
        & $\checkmark$ \\
        \bottomrule
    \end{tabular*}

    \vspace{2pt}
\end{table*}

\paragraph{Training Objectives}
We jointly optimize the instance segmentation, tracking, Gaussian reconstruction, and motion estimation modules while keeping the pretrained backbone frozen to preserve its visual and geometric priors.
The photometric loss $\mathcal{L}_{\mathrm{photo}}$ supervises RGB renderings of source and target frames, while $\mathcal{L}_{\mathrm{ins}}$ supervises instance classes and masks only in officially annotated source frames.
Nevertheless, incorrect motion predictions may place instance Gaussians behind the target camera, where they receive no photometric gradients from that view.
To provide a corrective signal, we introduce a smooth depth validity loss based on normalized camera-space depth as follows:
\begin{equation}
    \mathcal{L}_{\mathrm{valid}}
    =
    \underset{(t,p,u)\in\Omega_{\mathrm{ins}}}
              {\operatorname{Avg}}
    \left[
        \delta\,\left[
        \operatorname{softplus}\!\left(
            \frac{z_{\min}-\widetilde{z}_{t,p\rightarrow u}}
                 {\delta}
        \right)\right]
    \right],
    \label{eq:depth_validity_loss}
\end{equation}
where $u$ indexes target frames and
$\widetilde{z}_{t,p\rightarrow u}
=z_{t,p\rightarrow u}/d_{t,p}^{\mathrm{ref}}$.
Here, $z_{t,p\rightarrow u}$ denotes the camera-space depth of Gaussian $G_{t,p}$ after alignment to target frame $u$.
The reference depth $d_{t,p}^{\mathrm{ref}}$ is the median valid predicted depth of the corresponding instance in source frame $t$.
The parameters $z_{\min}$ and $\delta$ specify the normalized depth margin and penalty smoothness, respectively, and $\operatorname{softplus}$ is a smooth approximation to ReLU.
The set $\Omega_{\mathrm{ins}}$ contains the valid triplets $(t,p,u)$ retained for supervision, linking source Gaussians to annotated target frames in which the same tracked instances are observed.
The operator $\operatorname{Avg}$ averages pixel losses within each source observation, then across observations and target instances.
This constraint supplies gradients for correcting invalid depths even when photometric supervision is unavailable.
The overall training objective can be expressed as follows:
\begin{equation}
    \mathcal{L}
    =
    \mathcal{L}_{\mathrm{photo}}
    +
    \mathcal{L}_{\mathrm{ins}}
    +
    0.02\mathcal{L}_{\mathrm{valid}}.
    \label{eq:training_objective}
\end{equation}


\subsection{Instance Asset Completion and Scene Editing}
\label{sec:completion}
For each selected instance, we collect its source Gaussians using the predicted masks and persistent identities. We then align their centers to a common reference time $\tau_r$ using the predicted displacements $\Delta\boldsymbol{\mu}_{t,p}(\tau_r)$. This alignment accounts for the object's motion when combining observations captured at different times.
For Gaussian completion, we adopt a pretrained TRELLIS-based completion prior~\cite{trellis,3dfixer}, conditioned on the aggregated geometry and associated image appearance. The observed geometry provides spatial constraints from the reconstructed instance, while the prior infers plausible geometry and appearance for unobserved surfaces. The resulting Gaussian asset is placed in the shared scene coordinate system at the instance's reference pose and scale. A user-specified rigid transformation modifies the asset's position and orientation. Objects can also be inserted or removed by adding or excluding the corresponding Gaussian sets. The edited assets are jointly rendered with the remaining scene Gaussians from the desired viewpoint, enabling controllable re-simulation.

\section{Experiments}
\label{sec:experiment}

\begin{figure*}[!tp]
    \centering
    \includegraphics[width=0.95\linewidth]{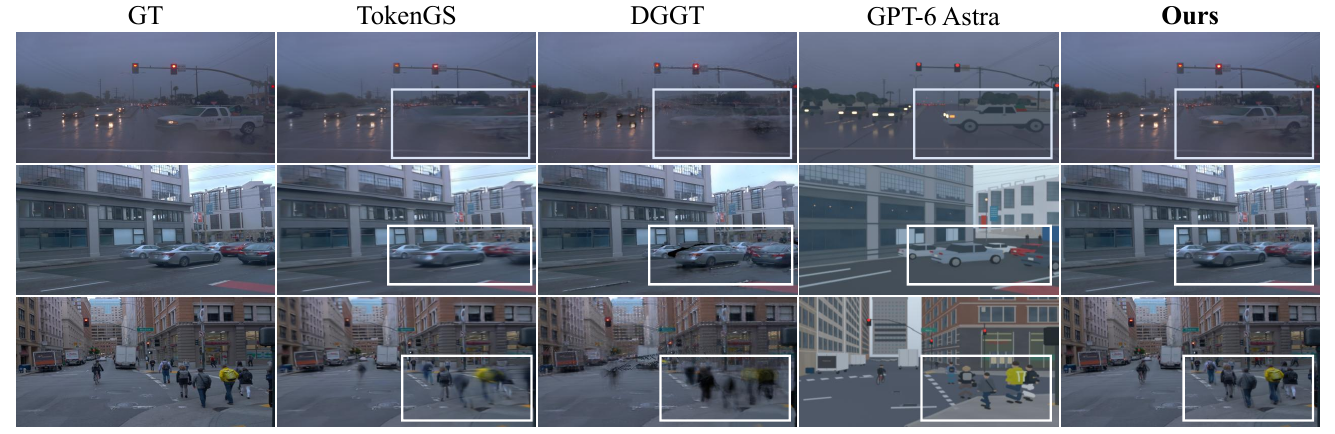}
    \caption{Qualitative comparison of different methods on the Waymo~\cite{waymo} dataset.}
    \label{fig:exp_vis_waymo}
\end{figure*}

\subsection{Experimental protocol}
\paragraph{Datasets}
We evaluate IGGLoop on the Waymo Open Dataset~\cite{waymo}, using 798 scenes for training and 202 held-out scenes for evaluation. These sequences capture diverse traffic interactions, lighting conditions, and weather conditions. To assess cross-dataset generalization, we further evaluate
the Waymo-trained model on the nuScenes~\cite{nuscenes} and Argoverse~2~\cite{argoverse2} datasets without fine-tuning, and conduct additional experiments with training on each target dataset. Following DGGT~\cite{dggt}, we measure rendering quality using PSNR, SSIM, and LPIPS, which assess pixel-level fidelity, structural similarity, and perceptual similarity, respectively. We additionally evaluate camera pose accuracy using AUC@3 and AUC@30, and assess global pointmap accuracy using relative L2 error (Rel. L2) and inlier ratio (IR) after Sim(3) alignment to the ground truth. We further evaluate instance motion using the 3D end-point error (EPE3D, in meters) and angular error (in radians).

\begin{table}[!tbp]
    \centering
    \caption{
        Quantitative comparison under zero-shot and trained
        settings on the nuScenes and Argoverse~2 datasets.
    }
    \label{tab:cross_dataset}
    \scriptsize
    \setlength{\tabcolsep}{1.75pt}
    \renewcommand{\arraystretch}{1.12}

    \begin{tabular*}{\columnwidth}{
        @{\extracolsep{\fill}}lccc|ccc@{}
    }
        \toprule
        \multirow{2}{*}{\textbf{Method}}
        & \multicolumn{3}{c|}{\textbf{nuScenes}}
        & \multicolumn{3}{c}{\textbf{Argoverse~2}} \\
        \cmidrule(lr){2-4}
        \cmidrule(lr){5-7}
        & \textbf{PSNR}$\,\uparrow$
        & \textbf{SSIM}$\,\uparrow$
        & \textbf{LPIPS}$\,\downarrow$
        & \textbf{PSNR}$\,\uparrow$
        & \textbf{SSIM}$\,\uparrow$
        & \textbf{LPIPS}$\,\downarrow$ \\
        \midrule

        \multicolumn{7}{@{}l}{\textbf{Zero-shot}} \\
        STORM~\cite{storm}
        & 24.09 & 0.689 & 0.23 &22.87 &0.712 &0.19 \\
        VGGT++~\cite{vggt}
        & 24.87 & 0.735 & 0.20 &24.33 &0.788 &0.14 \\
        ReconDrive~\cite{ReconDrive}
        & 23.63 & 0.663 & 0.30 &22.08 &0.664 &0.19 \\
        TokenGS~\cite{tokenGS}
        & \underline{26.50} & \underline{0.781} & 0.27 &\underline{25.52} &\underline{0.836} &0.20 \\
        Envision4D~\cite{envision4D}
        & 22.62 & 0.607 & 0.20 &23.69 &0.708 &\underline{0.12} \\
        DGGT~\cite{dggt}
        & 26.17 & 0.771 & \underline{0.16} &24.96 &0.791 &0.16 \\
        \textbf{Ours}
        & \textbf{27.27} & \textbf{0.807} & \textbf{0.14}
        &\textbf{26.25} &\textbf{0.848} &\textbf{0.11} \\
        \midrule

        \multicolumn{7}{@{}l}{\textbf{Trained}} \\
        DGGT~\cite{dggt}
        &27.43 &0.796 &0.13 &25.88 &0.812 &0.13 \\
        Ours
        &\textbf{28.49} &\textbf{0.822} &\textbf{0.11} &\textbf{27.11} &\textbf{0.859} &\textbf{0.10} \\
        \bottomrule
    \end{tabular*}
\end{table}

\subsection{Comparison with state-of-the-art methods}
\begin{table*}[!t]
    \centering
    \captionsetup{position=top,skip=4pt}
    \begin{minipage}[t]{0.56\textwidth}
        \vspace{0pt}
        \centering
        \caption{\textbf{Ablation on the number of Looped Transformer iterations $L$.}
        DGGT is included as an external reference.
        Global pointmap metrics are evaluated after Sim(3) alignment to the ground truth. Inference time is reported in seconds.}
        \label{tab:ablation_loop_iterations}

        \scriptsize
        \setlength{\tabcolsep}{1.1pt}
        \renewcommand{\arraystretch}{1.12}
        \begin{tabular*}{\linewidth}{@{\extracolsep{\fill}}lcccccccc@{}}
            \toprule
            \multirow{2}{*}{Method}
            & \multirow{2}{*}{$L$}
            & \multicolumn{2}{c}{Rendering}
            & \multicolumn{2}{c}{Camera pose}
            & \multicolumn{2}{c}{Pointmap}
            & \multirow{2}{*}{\shortstack{Time $\downarrow$}} \\
            \cmidrule(lr){3-4} \cmidrule(lr){5-6} \cmidrule(lr){7-8}
            & & PSNR $\uparrow$ & SSIM $\uparrow$
            & AUC@3 $\uparrow$ & AUC@30 $\uparrow$
            & Rel.\ L2 $\downarrow$ & IR $\uparrow$ & \\
            \midrule
            DGGT~\cite{dggt}
            & & 27.24 & 0.796 & 36.48 & 76.72 & 0.08 & 33.30 & 1.38 \\
            \midrule
            \multirow{4}{*}{Ours}
            & 4 & 27.07 & 0.799 & 27.88 & 76.66 & 0.08 & 32.54 & \textbf{0.30} \\
            & 8 & 27.64 & 0.820 & 59.94 & 83.57
            & \underline{0.07} & \underline{49.31} & \underline{0.34} \\
            & 16 & \textbf{27.94} & \textbf{0.826}
            & \textbf{66.66} & \textbf{84.40}
            & \textbf{0.07} & \textbf{50.86} & 0.40 \\
            & 32 & \underline{27.72} & \underline{0.825}
            & \underline{62.85} & \underline{83.80}
            & 0.07 & 48.72 & 0.53 \\
            \bottomrule
        \end{tabular*}
    \end{minipage}\hfill
    \begin{minipage}[t]{0.42\textwidth}
        \vspace{0pt}
        \centering
        \caption{\textbf{Instance-context ablation.}
        G/M denote Gaussian/motion context. EPE3D and angular error are in meters and radians. Other methods are external references.}
        \label{tab:context_ablation}

        \scriptsize
        \setlength{\tabcolsep}{1.1pt}
        \renewcommand{\arraystretch}{1.12}
        \begin{tabular*}{\linewidth}{@{\extracolsep{\fill}}lcccccc@{}}
            \toprule
            \multirow{2}{*}{Method}
            & \multicolumn{2}{c}{Context}
            & \multicolumn{2}{c}{Instance recon.}
            & \multicolumn{2}{c}{Instance motion} \\
            \cmidrule(lr){2-3} \cmidrule(lr){4-5} \cmidrule(lr){6-7}
            & G & M & PSNR $\uparrow$ & SSIM $\uparrow$
            & EPE3D $\downarrow$ & Angular $\downarrow$ \\
            \midrule
            DGGT~\cite{dggt}
            & -- & -- & 23.04 & 0.687 & 0.62 & 1.03 \\
            TokenGS~\cite{tokenGS}
            & -- & -- & 23.28 & 0.690 & -- & -- \\
            DynamicVGGT~\cite{dynamicvggt}
            & -- & -- & 18.41 & 0.565 & 1.72 & 1.37 \\
            \midrule
            \multirow{4}{*}{\textbf{IGGLoop}}
            & $\times$ & $\times$
            & 23.26 & 0.715 & 0.69 & 1.08 \\
            & $\times$ & $\checkmark$
            & 23.45 & 0.722 & \underline{0.48} & \underline{0.72} \\
            & $\checkmark$ & $\times$
            & \underline{24.79} & \underline{0.730} & 0.67 & 1.01 \\
            \cmidrule(lr){2-7}
            & $\checkmark$ & $\checkmark$
            & \textbf{25.07} & \textbf{0.748}
            & \textbf{0.44} & \textbf{0.70} \\
            \bottomrule
        \end{tabular*}
    \end{minipage}
\end{table*}

\begin{figure*}[!tp]
    \centering
    \includegraphics[width=0.80\linewidth]{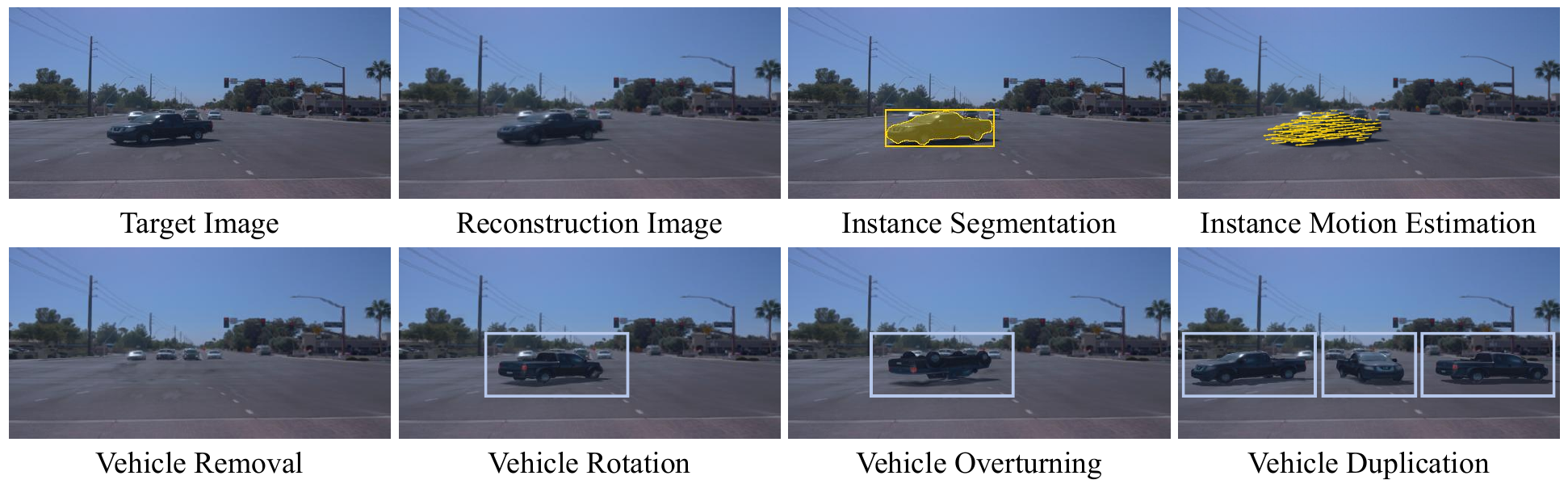}
    \caption{Scene editing results on the Waymo~\cite{waymo} dataset.}
    \label{fig:scene_editing}
\end{figure*}

To evaluate the effectiveness of the instance-grounded 4D Gaussian reconstruction framework, we compare IGGLoop with both per-scene optimization methods~\cite{3dgs,deformablegs} and recent feed-forward approaches~\cite{gs_lrm,noposplat,storm,vggt,ReconDrive,tokenGS,dynamicvggt,envision4D,dggt}. Additionally, we evaluate GPT-6 Astra~\cite{gpt6} as an MLLM-based agent baseline, given its reported capabilities in spatial understanding and 3D modeling.

To assess both reconstruction on input frames and novel view synthesis on interpolated frames, we use 10-frame, three-view sequences from scenes as the test set, taking 4 frames as input to predict the intermediate frames and evaluating performance on all frames. As reported in Table~\ref{tab:waymo_comparison}, IGGLoop achieves the best performance across all three evaluated metrics, with a PSNR of 27.94 dB. Notably, compared with the SOTA pose-free feed-forward reconstruction method DGGT, IGGLoop achieves these gains using only 6.4\% of its parameters (2752.94M vs.\ 176.52M), while delivering a $3.45\times$ speedup (1.38 s vs.\ 0.40 s). This parameter efficiency stems from our Looped Transformer design, which iteratively reuses the same attention blocks with shared weights instead of stacking independently parameterized layers. The speedup can be attributed in part to reduced reliance on off-the-shelf models for video segmentation and 3D point tracking. Fig.~\ref{fig:exp_vis_waymo} further provides qualitative comparisons with several representative approaches. Compared with other feed-forward methods, IGGLoop produces more faithful instance reconstructions. While GPT-6 Astra~\cite{gpt6} demonstrates promising spatial understanding by capturing the overall scene layout, it still struggles to produce photorealistic object reconstructions, particularly in recovering fine textures and appearance details. Additionally, to further assess cross-dataset generalization, we evaluate IGGLoop on nuScenes~\cite{nuscenes} and Argoverse~2~\cite{argoverse2} under both zero-shot and fine-tuning settings, as reported in Table~\ref{tab:cross_dataset}. IGGLoop also achieves the best performance across all three metrics. Additional qualitative results on these two datasets are presented in the supplement.

\subsection{Ablation studies and application}
\paragraph{Number of Looped Transformer Iterations}
To assess the impact of iterative refinement on reconstruction quality, geometric accuracy, and computational cost, we vary the number of Looped Transformer iterations. As shown in Table~\ref{tab:ablation_loop_iterations}, increasing $L$ from 4 to 16 improves PSNR by 0.87 dB, alongside gains in SSIM, camera pose accuracy, and pointmap quality, supporting the effectiveness of iterative geometric refinement. Further increasing $L$ to 32 adds computational cost without further gains in reconstruction or geometric accuracy, suggesting diminishing returns from additional refinement. 

\paragraph{Instance-Grounded Gaussian and Motion Prediction}
Table~\ref{tab:context_ablation} evaluates instance-region reconstruction and dynamic instance motion under four Gaussian and motion branch configurations to assess the contribution of the instance-grounded prediction paradigm. Without instance context, each branch directly decodes per-pixel predictions from global features, as in DGGT. Even with this configuration, IGGLoop outperforms DGGT in instance reconstruction, which we attribute in part to the Looped Transformer's camera pose and depth estimation capabilities~\cite{deja_view}, consistent with Table~\ref{tab:ablation_loop_iterations}. Adding Gaussian instance context improves PSNR by 1.53 dB, supporting cross-frame aggregation for object geometry and appearance. Motion instance context reduces EPE3D and further improves reconstruction performance, highlighting the connection between temporal alignment and rendering quality. Combining both achieves the best results across all metrics. These complementary gains support organizing observations by persistent instance identity for joint reconstruction and motion estimation. Ablations on adaptive fusion bandwidth and depth validity loss are provided in the supplement.

\paragraph{Instance-Grounded Scene Editing}
We demonstrate instance-grounded scene editing on Waymo in Fig.~\ref{fig:scene_editing}. The top row presents the reconstructed scene alongside instance segmentation and motion estimates. These predictions allow the selected vehicle to be extracted and its observations to be aligned for completion, following Sec.~\ref{sec:completion}. The bottom row shows vehicle removal, rotation, overturning, and duplication. Removal and duplication modify the presence and number of vehicles in the scene, while rotation and overturning change the orientation of the completed asset. In particular, these orientation changes expose surfaces absent from the original view, illustrating the role of completion in supporting edits beyond the observed pose. The rendered examples demonstrate that the reconstructed vehicle can be individually manipulated and reused within the scene, connecting instance-grounded reconstruction with controllable editing of driving environments.

\section{Conclusion}
\label{sec:conclusion}
This work demonstrates the value of organizing cross-frame observations around persistent instances for dynamic scene reconstruction. In the proposed framework IGGLoop, instance associations enable complementary observations to inform object appearance and temporal alignment, while a compact looped backbone supports geometric refinement through shared parameters. Beyond rendering, persistent identities and estimated motion further support observation aggregation and alignment for asset completion and editing. The resulting representation supports both image synthesis and the use of individual objects as editable assets for controllable driving simulation, highlighting the broader value of maintaining instance structure.

\clearpage
\section*{LLM Usage Statement}
We used ChatGPT and Codex (OpenAI) as assistive tools throughout manuscript preparation, including drafting and restructuring text, refining technical descriptions, improving language, and preparing LaTeX tables and document layout.
For the research, GPT-6 Astra was evaluated as an MLLM-based reconstruction baseline (Sec.~\ref{sec:experiment}).
The authors take full responsibility for the methodology, experimental results, interpretations, and final manuscript.
{
    \small
    \bibliographystyle{ieeenat_fullname}
    \bibliography{main}
}

\clearpage
\maketitlesupplementary
\raggedbottom
\setlength{\dbltextfloatsep}{10pt plus 2pt minus 2pt}
\captionsetup[figure]{skip=6pt}
\captionsetup[table]{skip=4pt}
\setlength{\textfloatsep}{12pt plus 2pt minus 2pt}

\section{Implementation Details}
\label{sec:implement}
\subsection{Baseline Selection}
We select representative baselines covering different reconstruction paradigms and scene representations. Among them, 3DGS~\cite{3dgs} and DeformableGS~\cite{deformablegs} provide per-scene optimization baselines with static and deformable Gaussian representations, respectively. GS-LRM~\cite{gs_lrm} and NoPoSplat~\cite{noposplat} are feed-forward Gaussian reconstruction methods, with NoPoSplat operating without input camera poses. STORM~\cite{storm}, TokenGS~\cite{tokenGS}, and Envision4D~\cite{envision4D} model scene dynamics through velocity prediction, time-conditioned Gaussian tokens, and nonlinear motion, respectively. ReconDrive~\cite{ReconDrive} incorporates instance masks and 3D box annotations for driving scene reconstruction, while VGGT++~\cite{vggt}, DynamicVGGT~\cite{dynamicvggt}, and DGGT~\cite{dggt} provide transformer-based, pose-free reconstruction baselines. We additionally include GPT-6 Astra~\cite{gpt6} as an MLLM-based agent baseline to assess its spatial reasoning and 3D modeling capabilities, using the \textit{Extra High} reasoning mode with the \textit{$1.5\times$} speed version.

\subsection{Task setup}
The experiments in Table~\ref{tab:waymo_comparison} and Fig.~\ref{fig:exp_vis_waymo} evaluate short-sequence reconstruction and prediction. Since existing methods adopt different sequence sampling strategies, we evaluate their publicly available implementations under a common sampling protocol for a fair comparison. For each of the 202 scenes in the Waymo Open Dataset~\cite{waymo} validation set, we select the first two 10-frame sequences, condition on frames with indices $0$, $3$, $6$, and $9$, and predict all ten frames. Inference is performed using the forward-facing, front-left, and front-right camera views. Evaluation covers both the observed input frames and the intermediate frames, jointly assessing basic scene reconstruction and novel view synthesis (NVS). Optimization-based methods are fitted to each selected sequence for 5,000 iterations, whereas feed-forward methods are evaluated without test-time optimization.

\subsection{Training and Evaluation Details}
\label{sec:training_details}
We initialize the DINOv2 encoder and Looped Transformer from the pretrained backbone~\cite{deja_view}, using 16 refinement iterations. The backbone, including its depth and ray heads, remains frozen, while the instance, Gaussian, and motion branches are jointly optimized. All images are resized to $512 \times 256$. Each training sample contains ten consecutive frames, with  all ten frames providing reconstruction supervision. Instance masks, class labels, and cross-frame associations used for supervision are obtained from Waymo's official annotations. Instance supervision mainly focuses on two categories, \textit{vehicle} and \textit{pedestrian}, with all vehicle categories in the annotations merged into a single \textit{vehicle} class. In addition to individual objects, our framework also identifies semantic regions such as sky and road, providing additional semantic information for subsequent scene editing. We train IGGLoop on the Waymo training split for approximately 32 hours using 8 NVIDIA A800 GPUs, with a batch size of 4 per GPU. We use AdamW with learning rates of $5 \times 10^{-5}$ for the instance branch and $1 \times 10^{-4}$ for the Gaussian and motion branches, and a weight decay of $1 \times 10^{-4}$. The learning rates follow a cosine decay schedule after 1,000 warm-up iterations. We employ bfloat16 mixed-precision training and clip the gradient norm to 1.0. We evaluate inference time on a single NVIDIA A800 GPU, performing warm-up runs before timing to reduce initialization overhead and obtain stable runtime measurements.

\section{Additional Experimental Results}
\label{sec:additional}
\begin{figure*}[!t]
    \centering
    \includegraphics[width=0.83\textwidth]{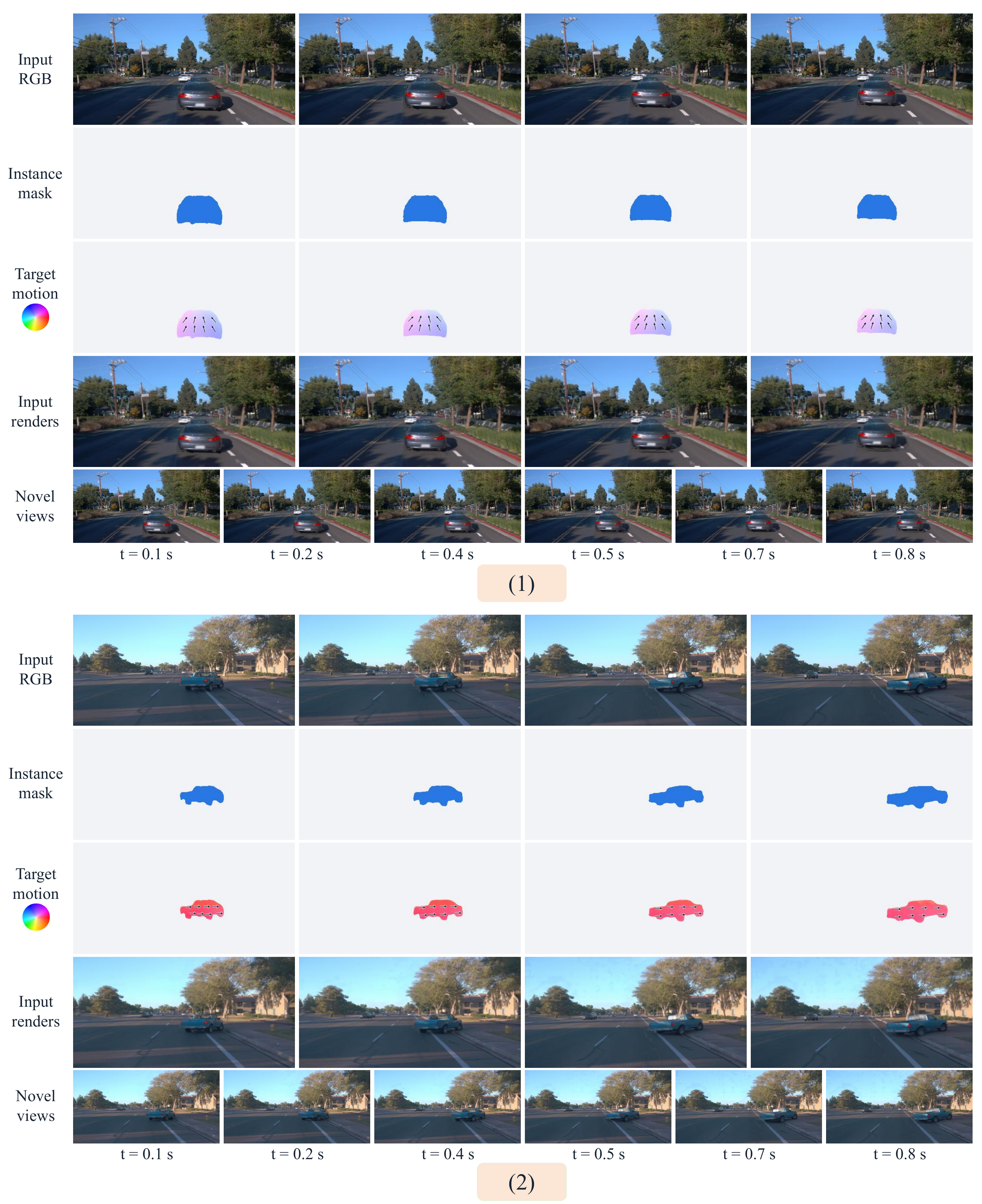}
    \caption{\textbf{Additional qualitative results on Waymo.}
    For each example, rows show input RGB images, predicted instance masks, target-conditioned motion, input-frame reconstructions, and novel-view renderings at intermediate timestamps.}
    \label{fig:add_vis}
\end{figure*}

\begin{figure*}[!t]
    \centering
    \includegraphics[width=0.83\textwidth]{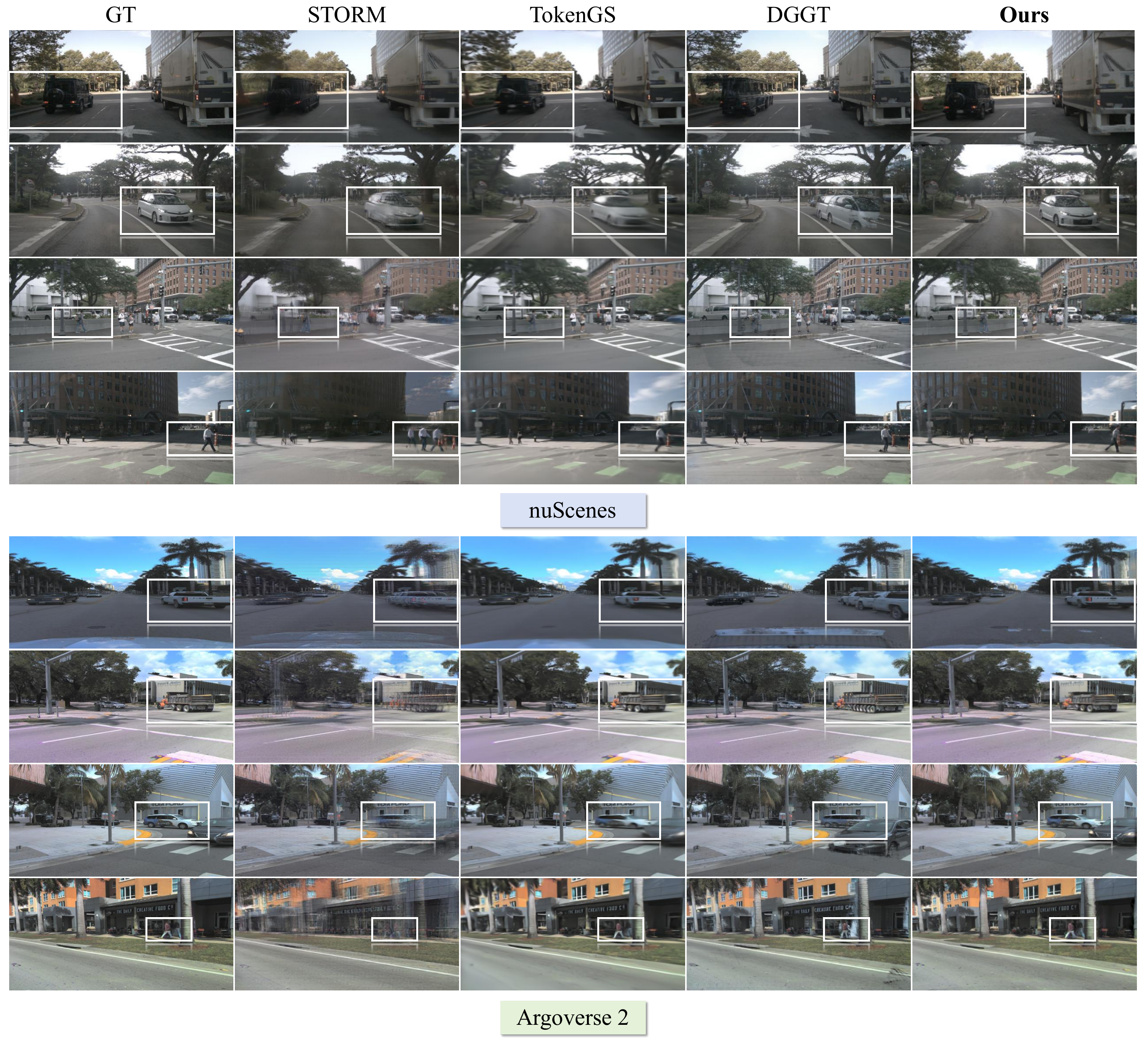}
    \caption{Qualitative comparison of different methods on the nuScenes and Argoverse~2 datasets.}
    \label{fig:exp_vis_zeroshot}
\end{figure*}

\subsection{Comparison on Additional Datasets}
\label{sec:additional_datasets}

We further evaluate IGGLoop on nuScenes~\cite{nuscenes} and Argoverse~2~\cite{argoverse2} under two settings: zero-shot evaluation and target-domain training.
In the zero-shot setting, we directly apply the source-domain checkpoint without fine-tuning on either target dataset, assessing its generalization to unseen driving environments.
The target-domain training setting additionally uses training data from the corresponding dataset to evaluate the benefits of domain-specific adaptation. 
Sampling and split details are as follows. For nuScenes
(v1.0), the dataset contains approximately 1,000 driving
scenes, each lasting roughly 20 s, with camera sampling
at 12 Hz. We use its official training split of 700 scenes for target-domain adaptation and evaluate on all 150 validation scenes. For the Argoverse~2 Sensor Dataset, which comprises roughly 1,000 annotated driving sequences and provides multi-modal sensor observations, the camera trigger frequency is approximately 20 Hz.
We use its official training split of 700 scenes for target-domain adaptation and evaluate on the first 100 validation scenes.

Across datasets, we follow the Waymo preprocessing setup and resize all camera images to $512 \times 256$ without cropping.
We retain the same model architecture with 16 Looped Transformer iterations and use consistent PSNR, SSIM, and LPIPS evaluation settings. For zero-shot evaluation, the source-domain checkpoint is applied directly without parameter updates.
Additional zero-shot qualitative comparisons on nuScenes and Argoverse~2 are presented in Fig.~\ref{fig:exp_vis_zeroshot}.

\subsection{Additional Ablation Studies}
\begin{table}[!tbp]
    \centering
    \caption{\textbf{Ablation on the depth validity loss and temporal fusion.}
    Each variant removes one component from the full model.}
    \label{tab:ablation_depth_validity}
    \footnotesize
    \setlength{\tabcolsep}{4pt}
    \renewcommand{\arraystretch}{1.12}
    \begin{tabular*}{\columnwidth}{@{\extracolsep{\fill}}lccc@{}}
        \toprule
        \textbf{Variant}
        & \textbf{PSNR}$\,\uparrow$
        & \textbf{SSIM}$\,\uparrow$
        & \textbf{LPIPS}$\,\downarrow$ \\
        \midrule
        w/o $\mathcal{L}_{\mathrm{valid}}$
        & 25.18 & 0.763 & 0.18 \\
        w/o bandwidth
        &27.31  &0.814  &0.13  \\
        \textbf{Ours}
        &\textbf{27.94} & \textbf{0.826} & \textbf{0.12} \\
        \bottomrule
    \end{tabular*}
\end{table}
\paragraph{Depth Validity Loss}
To evaluate the contribution of the depth validity loss, we compare reconstruction performance with and without this term in Table~\ref{tab:ablation_depth_validity}. Adding the loss substantially improves reconstruction quality across all three metrics, which suggests that image reconstruction supervision benefits from an explicit constraint on target-view Gaussian depths. When predicted displacements move instance Gaussians behind the target camera, the photometric objective provides no corrective gradients from that view. The depth validity loss addresses this limitation by penalizing invalid depths, providing a complementary signal for motion learning. 

\paragraph{Adaptive Fusion Bandwidth}
Motion estimation errors can leave source Gaussians misaligned at the target time. The fusion head draws on the same local motion features, instance context, and temporal encoding as the motion head to predict adaptive bandwidths. By modulating Gaussian opacities, the resulting fusion weights can downweight Gaussians associated with unreliable motion estimates, reducing their influence on the rendered image. As shown in Table~\ref{tab:ablation_depth_validity}, removing bandwidth-based weighting degrades reconstruction quality. These results suggest that adaptive temporal fusion helps mitigate the effects of imperfect motion alignment when combining source observations.

\subsection{Additional Qualitative Results}
Fig.~\ref{fig:add_vis} presents additional qualitative results on Waymo, including input RGB images, predicted instance masks, target-conditioned motion, input-frame renderings, and novel views at intermediate timestamps.
The two examples show a vehicle ahead of the ego vehicle and a vehicle crossing an intersection, respectively.
Across source frames, the predicted masks consistently localize the selected vehicle despite changes in its appearance and image position, while the motion predictions provide instance-specific information for temporal alignment.
The input-frame renderings preserve the overall scene structure and vehicle appearance, and the intermediate novel views capture changes in vehicle position while maintaining coherent appearance.
These examples illustrate how instance associations support joint Gaussian reconstruction and motion estimation in different driving scenarios.
By propagating instance queries across frames and aggregating instance-specific observations, IGGLoop can also accommodate more input frames and longer video sequences. We further demonstrate this capability through additional visualizations in the accompanying demo.


\end{document}